\documentclass[preprint,10pt]{elsarticle}
\usepackage{setspace}
\usepackage{geometry}
\usepackage{graphicx}
\usepackage{physics}
\usepackage{amsmath}
\usepackage{tikz}
\usepackage{mathdots}
\usepackage{yhmath}
\usepackage{cancel}
\usepackage{color}
\usepackage{siunitx}
\usepackage{array}
\usepackage{multirow}
\usepackage{amssymb}
\usepackage{gensymb}
\usepackage{tabularx}
\usepackage{extarrows}
\usepackage{booktabs}
\usetikzlibrary{fadings}
\usetikzlibrary{patterns}
\usetikzlibrary{shadows.blur}
\usetikzlibrary{shapes}
\usepackage{diagbox}
\usepackage{bbding}
\usepackage{amsthm}
\usepackage{float}
\usepackage{bm}
\usepackage{pdflscape}
\usepackage{rotating}
\usepackage{booktabs}

\theoremstyle{definition}

\journal{Information Fusion}

\begin{document}

\begin{frontmatter}

\title{Domain-Adversarial Anatomical Graph Networks \\ for Cross-User Human Activity Recognition}

\author[inst1]{Xiaozhou Ye\corref{cor1}\fnref{fn1}}
\ead{xye685@aucklanduni.ac.nz}

\author[inst1]{Kevin I-Kai Wang\corref{cor2}}
\ead{kevin.wang@auckland.ac.nz}
\cortext[cor2]{Corresponding author}

\affiliation[inst1]{organization={Department of Electrical, Computer, and Software Engineering, The University of Auckland}, city={Auckland}, country={New Zealand}}

\begin{abstract}

Cross-user variability in Human Activity Recognition (HAR) remains a critical challenge due to differences in sensor placement, body dynamics, and behavioral patterns. Traditional methods often fail to capture biomechanical invariants that persist across users, limiting their generalization capability. We propose an Edge-Enhanced Graph-Based Adversarial Domain Generalization (EEG-ADG) framework that integrates anatomical correlation knowledge into a unified graph neural network (GNN) architecture. By modeling three biomechanically motivated relationships together—Interconnected Units, Analogous Units, and Lateral Units—our method encodes domain-invariant features while addressing user-specific variability through Variational Edge Feature Extractor. A Gradient Reversal Layer (GRL) enforces adversarial domain generalization, ensuring robustness to unseen users. Extensive experiments on OPPORTUNITY and DSADS datasets demonstrate state-of-the-art performance. Our work bridges biomechanical principles with graph-based adversarial learning by integrating information fusion techniques. This fusion of information underpins our unified and generalized model for cross-user HAR.
\end{abstract}

\begin{keyword}
Human activity recognition \sep deep domain generalization \sep edge-enhanced graph neural networks \sep adversarial learning

\end{keyword}

\end{frontmatter}

\section{Introduction}

Human Activity Recognition (HAR) using wearable sensors has transformative applications in healthcare, sports, and smart environments. However, deploying HAR systems across diverse users faces a fundamental challenge: cross-user variability. Differences in body morphology (e.g., limb length, muscle mass) and movement styles (e.g., gait patterns) lead to significant distribution shifts in sensor data. For instance, accelerometer readings from a wrist sensor during "drinking from a cup" can vary substantially between users due to differences in arm motion and grip style. Traditional machine learning models, which assume identical training and testing distributions, often fail to generalize under such shifts.

Conventional HAR methods typically involve feature extraction followed by classification using models such as Convolutional Neural Networks (CNNs) or Recurrent Neural Networks (RNNs)~\cite{hammerla2016deep,ordonez2016deep}. Although these methods are effective within individual datasets, they inherently learn user-specific patterns, leading to significant performance degradation when applied to unseen users~\cite{hammerla2016deep}. To address this limitation, recent research has explored domain adaptation and transfer learning techniques~\cite{wang2018deep}. However, these approaches rely heavily on labeled target-user data, which is often impractical to obtain in real-world scenarios.

Domain generalization offers a promising alternative by handling scenarios where no data from the target user(s) is available during training~\cite{muandet2013domain,li2018learning}. Despite its potential, most current domain generalization methods focus primarily on aligning user-specific features without considering shared biomechanical patterns that persist across users~\cite{ahmad2021graph}. Even though users differ in attributes like gender, weight, and height, certain anatomical correlations between body parts remain consistent across individuals.

Human anatomy imposes shared biomechanical constraints that govern movement patterns, which can serve as a powerful prior knowledge for HAR models. For example, interconnected units, such as adjacent body parts like the upper and lower arm, exhibit coupled motion during activities like "opening a door". Similarly, analogous units, including bilateral limbs like the left and right knees, tend to synchronize during symmetrical activities such as "jumping". Additionally, lateral units, which involve body parts on the same side, such as the left arm and left leg, coordinate during lateral movements like "side lunges". These biomechanical invariants persist across users and represent anatomical correlations between sensor positions. Despite their significance, existing domain generalization methods often overlook this structural knowledge, focusing solely on feature alignment without leveraging these intrinsic relationships. Incorporating such correlations into HAR models has the potential to enhance their generalization capabilities across diverse users.

Despite the promise of leveraging anatomical correlation knowledge, several technical challenges remain. One challenge is the heterogeneous sensor relationships, as different body parts exhibit varying degrees of correlation depending on the activity being performed; for example, while leg movements dominate activities like "running", arm movements are more prominent in tasks such as "throwing". Additionally, domain-invariant representation learning poses a significant challenge, as it requires extracting activity-relevant features while effectively disentangling these features from user-specific biases to ensure robust generalization across diverse users. Addressing these challenges is critical to building effective cross-user HAR models.

To address these challenges, we propose the \textbf{Edge-Enhanced Graph-Based Adversarial Domain Generalization (EEG-ADG)} framework, which integrates anatomical correlation knowledge into a unified Graph Neural Network (GNN) architecture. Our key contributions include:
\begin{itemize}
    \item \textbf{Unified Graph Representation of Multi-Type Anatomical Correlations:} Unlike methods that treat different anatomical correlations (Interconnected, Analogous, and Lateral Units) separately, our approach constructs a single unified graph. This integration allows the model to capture the simultaneous influence of spatial, functional, and positional relationships, which is essential for representing the complex and common coordination involved in human activities among users.
    \item \textbf{Dynamic and Adaptive Edge-Enahnced GNN architecture:} An novel Variational Edge Feature Extractor technique is proposed for capturing variability in edge features, enabling the model to learn robust and generalizable representations of inter-sensor relationships. In addition, an adaptive edge attention mechanism is proposed that dynamically weights the importance of inter-sensor relationships, allowing the model to focus on the most relevant correlations tailored to each activity. This adaptive treatment of edge information enables the model to capture higher-order, context-sensitive interactions that are critical for accurate cross-user activity recognition.
    \item \textbf{Adversarial Domain Generalization for User-Invariant Feature Learning:} By incorporating an adversarial training strategy, our model is designed to learn features that are invariant across different users. This addresses the common issue of cross-user variability in HAR tasks, ensuring that the learned representations are robust and can generalize well to unseen users.
\end{itemize}

We evaluate EEG-ADG on two benchmark datasets, \textit{OPPORTUNITY} and \textit{DSADS}, and demonstrate its state-of-the-art performance in cross-user HAR tasks. Our framework fuses biomechanical principles with graph-based adversarial learning, offering a robust solution to the cross-user variability problem in HAR systems.

The rest of the paper is organized as follows. Section~\ref{sec:related_work} provides background on cross-user variability and GNN in HAR. Section~\ref{sec:method} details our proposed EEG-ADG method. Section~\ref{sec:experimental_setup} describes the experimental setup and results, and Section~\ref{sec:conclusion} concludes the paper and outlines future work.

\section{Related Work}
\label{sec:related_work}

\subsection{Cross-User Variability in HAR}

Cross-user variability in HAR arises from differences in sensor placement \cite{gil2023reducing}, body dynamics, and behavioral patterns \cite{soleimani2021cross}. These differences lead to distribution shifts that challenge traditional machine learning models \cite{wang2024optimization}, particularly in critical applications such as healthcare, surveillance, and smart environments, where robust performance across diverse users is essential \cite{gupta2022human}. To address this issue, HAR research has proposed several methods, which can be broadly categorized into fine-tuning \cite{guo2021efficacy}, domain adaptation \cite{ye2023cross}, and domain generalization \cite{bento2023exploring}.

\textbf{Fine-tuning} involves adapting a pre-trained model to a specific dataset, enhancing its performance for particular users or contexts \cite{thukral2025cross}. Ye et al. \cite{ye2024deep} highlight how distribution shifts in HAR data challenge traditional models, emphasizing the need for adaptive techniques. Manuel et al. \cite{gil2023reducing} note performance drops when models are tested on unseen users, underscoring the potential of fine-tuning. Recent studies further explore this approach. Gopalakrishnan et al. \cite{gopalakrishnan2024comparative} compare fine-tuning deep networks for action recognition, demonstrating improved accuracy in surveillance scenario. Genc et al. \cite{genc2024human} leverage fine-tuned CNN-LSTM models to enhance activity recognition from sensor data. Leite et al. \cite{leite2022resource} explore combining continual learning with fine-tuning to address dynamic data activity classification scenarios. These advancements highlight the efficacy of fine-tuning but also reveal scalability and privacy challenges due to the need for user-specific data, with the limitation of requiring labeled target user data, which is often difficult to access in real-world scenarios \cite{YeSensor2024}.

\textbf{Domain adaptation} only requires the unlabeled target user data \cite{ye2025cross}. It adjusts models trained on source domain to perform well on target domain, addressing cross-user variability by aligning source and target data distributions \cite{xiaozhou2024genDATR}. Avijoy et al. \cite{chakma2021activity} leverages adversarial learning to select the most relevant feature representations from multiple source domains and map them to the target domain by learning perplexity scores. These scores reflect the relevance of each source domain to the target domain, allowing the model to prioritize the most useful features. Hu et al. \cite{hu2023swl} introduce sample weight learning, where each sample is assigned a weight based on its classification loss and domain discrimination loss. These weights are calculated using a parameterized network, which is optimized end-to-end through a meta-optimization update rule. This rule is guided by the meta-classification loss on pseudo-labeled target samples, allowing the model to adaptively learn a weighting function tailored to the specific cross-user HAR task. Napoli et al. \cite{napoli2024benchmark} introduce a benchmark for evaluating domain adaptation and generalization in smartphone-based HAR. These works underscore the potential of domain adaptation but highlight its dependency on target data as a limitation.

\textbf{Domain generalization} aims to train models that can generalize to unseen domains without requiring target data, addressing a critical gap in real-world HAR applications \cite{hong2024crosshar}. For example, Bento et al. \cite{bento2023exploring} investigate regularization methods like Mixup and Sharpness-Aware Minimization for domain generalization in accelerometer-based HAR. Data augmentation is an important approach in domain generalization \cite{li2021simple}. By artificially increasing the variability in the training data, models can become more robust to different domains, Volpi et al. \cite{volpi2018generalizing} demonstrate generalizing to unseen domains via adversarial data augmentation that helps models learn to handle diverse user-specific variations.

In domain generalization, a prominent approach involves invariant feature learning, which emphasizes capturing characteristics that are stable and consistent across different domains \cite{lu2022domaininvariant}. By emphasizing these invariant features, models can better generalize to new users without overfitting to user-specific characteristics. Several advanced techniques have been developed for domain generalization. ANDMask \cite{parascandolo2021learning} enforces updates for invariant features across domains while preventing the model from memorizing domain-specific features. AdaRNN \cite{du2021adarnn} addresses temporal covariate shifts in time series data. By using adaptive learning techniques, it matches distributions across domains, ensuring robust generalization even when temporal patterns vary between users. DIFEX \cite{lu2022domaininvariant} combines internal and mutual invariance to enhance feature diversity, addressing the complex cross-domain needs of HAR. It leverages consistency within and across domains to learn robust features that capture a wide range of activity patterns. RSC \cite{huang2020self} discards dominant features during training, forcing the model to rely on label-related features that are more likely to be invariant across domains.

However, current domain generalization approaches are limited by their difficulty in capturing all relevant invariances across diverse users and their reliance on the diversity of source domains, which may not fully represent the variability in unseen target domains. Therefore, there is a critical need for innovative methods that can uncover the common, essential patterns in human activities, such as the consistent coordination of body parts during activities. These methods should focus on mining characteristics that are shared across individuals, enabling robust generalization even with limited source domain diversity.

\subsection{Graph Neural Networks for HAR}

GNNs have emerged as a powerful framework for modeling relational and structured data, distinguishing themselves from traditional neural networks by their ability to capture complex dependencies between entities \cite{wu2020comprehensive}. In GNNs, entities are represented as nodes, and their relationships as edges, forming a graph structure that encodes both spatial and functional interconnections \cite{seo2018structured}. This capability is particularly relevant to HAR, where activities involve consistent coordination of body parts across individuals—a key invariance that domain generalization methods must uncover. In HAR, body parts can be modeled as nodes, with edges representing spatial relationships (e.g., adjacency of limbs) or functional dependencies (e.g., correlated movements), offering a natural framework to mine shared activity patterns.

The usage of GNNs in HAR has shown promise, particularly in skeleton-based tasks, where inter-part dependencies are critical for recognizing actions. For example, Si et al. \cite{si2019attention} introduced an attention-enhanced GNN for skeleton-based action recognition, highlighting the importance of simultaneously modeling spatial and temporal dynamics. Similarly, Shi et al. \cite{shi2019two} extended this with two-stream adaptive GNNs, improving HAR by modeling body inter-part dependencies. Wieland et al. \cite{wieland2023tinygraphhar} proposed TinyGraphHAR demonstrating GNNs' discriminative power for complex activities. These studies highlight GNNs' strength in capturing the inter-part coordination. However, there has been limited exploration of GNNs in domain generalization. This gap is notable because consistent inter-part coordination across users can serve as a critical enabler for improved generalization, especially when the diversity of source users is constrained. Leveraging this consistency could provide a robust foundation for developing domain-invariant representations.

Moreover, GNNs' potential to sensor-based HAR remains underexplored. Unlike skeleton-based HAR, sensor-based scenarios involve dynamic inter-sensor relationships influenced by varying users and activity-specific patterns. Traditional GNNs, which often rely on static or predefined edge features \cite{avelar2020superpixel}, struggle to adapt to these evolving dependencies, limiting their effectiveness in cross-user settings. This gap highlights the need of methods that can uncover shared, essential patterns in sensor data, robustly generalizing across diverse users.

To address this, edge-enhanced GNNs is a promising extension on conventional GNNs, incorporating learnable, dynamic edge features that adapt to activity- and user-specific variations \cite{yan2024graph}. By modeling inter-sensor relationships more flexibly, these GNNs can capture common coordination patterns—such as those between sensors on different body parts—while accommodating individual differences. This adaptability makes edge-enhanced GNNs particularly suited for cross-user sensor-based HAR, where static models fall short, offering a pathway to improved generalization without requiring extensive source domain diversity.

In summary, GNNs excel at modeling spatial dependencies in HAR, yet their potential for domain generalization—especially in sensor-based, cross-user contexts—remains underexplored. This paper leverages edge-enhanced GNNs to address this gap, focusing on dynamic inter-sensor relationships to uncover essential, shared activity patterns. By doing so, it advances the development of robust, generalizable HAR models capable of overcoming the limitations of current domain generalization approaches.

\section{Method}
\label{sec:method}

\subsection{Problem Formulation}

In the cross-user HAR problem under the domain generalization setting, we have labeled data from \( K \) source users. The dataset for each source user \( k \) is represented as 
\[
S_k^{\text{Source}} = \left\{ \left( x_i^k, y_i^k \right) \right\}_{i=1}^{n_k},
\]
where \( n_k \) denotes the number of data instances for user \( k \), \( x_i^k \) represents the feature vector, and \( y_i^k \) represents the corresponding activity label. Each source user \( k \) is associated with a unique data distribution \( P_k^{\text{Source}} \), and these distributions differ across users, i.e., \( P_i^{\text{Source}} \neq P_j^{\text{Source}} \) for \( i \neq j \), where \( i, j \in \{1, 2, \dots, K\} \). 

In addition to the labeled source datasets, there are \( M \) unseen target users whose data is represented as 
\[
S_m^{\text{Target}} = \left\{ x_i^m \right\}_{i=1}^{n_m},
\]
where \( x_i^m \) denotes the feature vectors of the target users \( m \). Each target user \( m \) has its own distinct data distribution \( P_m^{\text{Target}} \), which differs not only from the distributions of other target users but also from those of the source users, i.e., \( P_k^{\text{Source}} \neq P_m^{\text{Target}} \) for all \( k \) and \( m \).

The feature space (i.e., the set of features describing the sensor data) and the label space (i.e., the set of activity classes) are consistent across all users. The objective of this problem is to train a model \( f \) using only the labeled data from the \( K \) source users such that it can accurately predict the activity labels \( \{ y_i^m \}_{i=1}^{n_m} \) for the \( M \) unseen target users, despite the distribution shifts between the source and target users.

\subsection{Edge-Enhanced Graph Neural Network with Adversarial Domain Generalization}

In this paper, we propose a \textbf{Sensor Position-Aware Edge-Enhanced Graph Neural Network with Adversarial Domain Generalization (EEG-ADG)} method. Our primary objective is to learn common, domain-invariant features across users by leveraging anatomical correlations that remain consistent despite cross-user variability. To achieve this, the model is designed to extract and generalize shared spatiotemporal patterns from sensor data, ensuring that features useful for recognizing coordinated human activities are not confounded by user-specific noise.

\begin{figure*}[h!]
\centering
\includegraphics[width=0.9\textwidth]{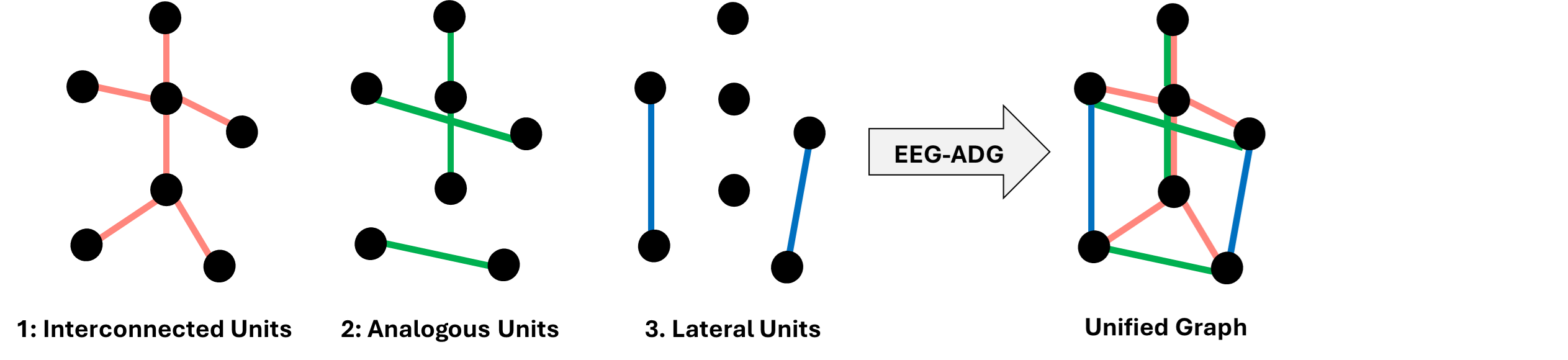}
\caption{Graph construction on three common Anatomical Units correlation.\label{3_categories_graphs}}
\end{figure*}

We unify three different types of Anatomical Units as shown in Figure~\ref{3_categories_graphs}: (1) Interconnected Units, which capture dependencies between adjacent sensors; (2) Analogous Units, which cluster sensors located on symmetrical or functionally related body parts; and (3) Lateral Units, which link sensors based on their spatial positioning to reflect coordination within specific regions. EEG-ADG aims to construct a unified graph where nodes represent sensor positions and edges correspond to all three anatomical correlations, enabling the model to integrate spatial, functional, and positional relationships simultaneously. This integration is crucial for capturing the higher-order interactions and nuanced patterns of movement that are common across different users.

Traditional graph neural networks leverage both node and edge data; however, they often rely on fixed or predetermined edge features that do not adapt to changes in sensor relationships. In contrast, our EEG-ADG framework is explicitly designed for domain generalization—it dynamically adjusts edge features so that the learned representations are not only robust for activity recognition, but also invariant across diverse user domains. This is achieved via the key component: Variational Edge Feature Extractor.

Instead of relying on static edge descriptors, Variational Edge Feature Extractor learns a probabilistic representation for each edge. This adaptive learning captures the variability in inter-sensor relationships while emphasizing features that are common across users, thus fostering domain invariance. Beyond incorporating edge features in GNN learning, Edge Attention Mechanism dynamically assigns importance weights to edges based on their context. By highlighting more informative connections and down-weighting noisy ones, the model focuses on shared anatomical correlations that remain consistent across different users, thereby reinforcing the domain generalization objective.

\begin{figure*}[h!] \centering \includegraphics[width=0.9\textwidth]{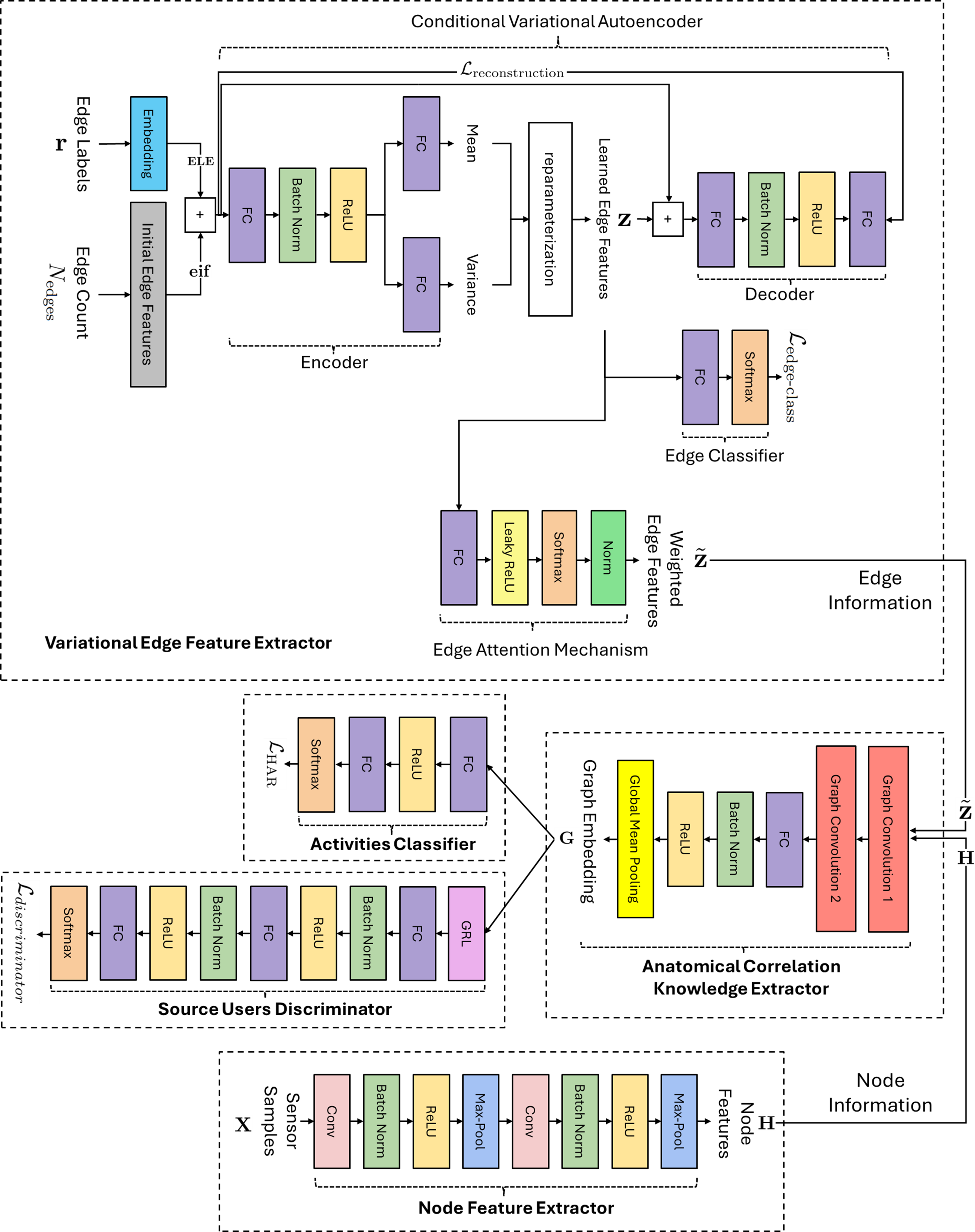} \caption{EEG-ADG framework overview.\label{GNN-ADG-framework}} \end{figure*}

Figure~\ref{GNN-ADG-framework} illustrates the overall EEG-ADG framework, which integrates multiple components: the Variational Edge Feature Extractor, the Node Feature Extractor, the Anatomical Correlation Knowledge Extractor, the Source Users Discriminator, and the Activities Classifier. In our design, every component is purposefully aligned with the core goal of learning user-invariant features. The adversarial domain generalization further ensures that the extracted features are common to all users, which enhances the model’s generalizability to unseen subjects.

These innovations collectively address cross-user variability by focusing on anatomical correlations that are stable across users and fusing multiple sensor information. By enforcing an adaptive and context-sensitive treatment of edge features, EEG-ADG not only captures the intricate interplay of sensor interactions but also ensures that the learned representations are robust and domain-invariant. This unified design provides a robust and scalable solution for cross-user domain generalization in HAR tasks.

\subsubsection{Variational Edge Feature Extractor}

To better learn and leverage the common edge features across users, the Variational Autoencoder (VAE) \cite{kingma2013auto} is applied to capture and utilize the three types of sensor positional biomechanical patterns shared across individuals. These anatomica patterns include Interconnected Units, Analogous Units, Lateral Units, which collectively represent consistent relationships among users. By abstracting these patterns into a latent representation, the VAE enables the model to generalize effectively across users \cite{ye2024deep}, focusing on underlying biomechanical structures rather than user-specific nuances as shown in Figure~\ref{GNN-ADG-framework} Variational Edge Feature Extractor component.

The probabilistic framework of the VAE is particularly well-suited for addressing the variability inherent in inter-sensor relationships across users. By introducing uncertainty into the representation of edge features, the VAE accounts for natural differences in how individuals perform activities or interact with sensors. This capacity to model variability ensures that the learned edge features are robust to changes in user behaviors, allowing the model to adapt to new or unseen users with greater flexibility.

A key advantage of the VAE lies in its ability to create a structured and smooth latent space. VAE regularizes the latent space, encouraging the model to capture essential biomechanical patterns while avoiding overfitting to individual-specific variations \cite{higgins2017beta}. This regularization helps the VAE distill meaningful, compact, and generalizable representations of edge features that align with the shared biomechanical structures observed across users.

By integrating these properties, the VAE excels at extracting user-invariant edge features. It effectively identifies shared biomechanical patterns, encodes variability and filters out noise or individual-specific deviations. As a result, the model becomes more robust, adaptable, and capable of generalizing across diverse user populations. This design ensures that the learned edge features are both scalable and effective for capturing the complex relations between sensors.

\paragraph{Initial Edge Features Generation}

A critical initial step is to generate robust and distinguishable edge features that balance structure and variability in the graph as shown in Figure~\ref{GNN-ADG-framework} Edge Initial Features component. By initializing base features with linearly spaced values, this approach ensures that edge features are structured and systematically distributed across a defined range, enabling the model to distinguish between different edges effectively. Adding Gaussian noise simulates real-world variability in sensor position relations or data collection environments, while maintaining the underlying structure provided by the base features. This design creates Edge Initial Features that are both distinguishable and robust.

The base features are created using a linearly spaced vector scaled by a factor $\alpha$. They are defined as:
\begin{equation}
\mathbf{b} = \alpha \cdot \text{linspace}(r_{\min},\, r_{\max},\, N_{\text{edges}}),
\end{equation}
where:
\begin{itemize}
    \item $r_{\min}$ and $r_{\max}$ define the range of the linear space
    \item $N_{\text{edges}}$ is the number of edges
    \item $\alpha$ is a scaling factor to ensure sufficient separation between the base features
\end{itemize}

Then, Gaussian noise is added to the base features to simulate real-world variability. The initial edge features are computed as:
\begin{equation}
\mathbf{eif} = \mathbf{b} + \beta \cdot \mathbf{n},
\end{equation}
where:
\begin{itemize}
    \item $\mathbf{n} \sim \mathcal{N}(0,1)$ represents standard Gaussian noise
    \item $\beta$ controls the magnitude of the noise
\end{itemize}

This formulation guarantees that the edge features $\mathbf{eif}$ preserve a structured base given by $\mathbf{b}$ while incorporating realistic variability through the noise term $\beta \cdot \mathbf{n}$.

\paragraph{Edge Label Embedding}

To represent edge labels \(\mathbf{r}\) (i.e. Anatomical Units types) as dense vectors, an embedding layer is used. Each label \(r \in \{0, 1, \dots, R-1\}\), where \(R\) is the total number of relation types, is mapped to a \(d\)-dimensional vector through the embedding matrix \(\mathbf{W} \in \mathbb{R}^{R \times d}\). Here, \(R\)=3 in our task, i.e. three types of anatomical correlation. Mathematically, the embedding is defined as:

\[
\mathbf{ELE}_r = \mathbf{W}[r]
\]

Here, \(\mathbf{ELE}_r \in \mathbb{R}^d\) is the embedding vector for the label \(r\). The embedding matrix \(\mathbf{W}\) is initialized randomly and learned during training. Therefore, the label embeddings \(\mathbf{ELE} = [\mathbf{ELE}_0, \mathbf{ELE}_1, \dots, \mathbf{ELE}_R-1] \in \mathbb{R}^{R \times d}\) form a matrix where each row corresponds to the embedding vector of a specific label.

\paragraph{Conditional Variational Autoencoder}
The goal of this component is to refine the edge features and extract a latent representation that emphasizes common biomechanical patterns. The embedded edge labels and the edge initial features are concatenated to form the input data for the VAE. The VAE employed is a Conditional VAE (CVAE), which incorporates an additional conditioning mechanism to guide the latent space. Specifically, an Edge Classifier is connected to the sampled latent space as shown in Figure~\ref{GNN-ADG-framework}, providing a supervisory signal to enforce structure and meaningful representation in the latent space.

The edge classifier is trained to predict the type of anatomical correlation for each edge. This encourages the CVAE to organize the latent space to reflect the semantic distinctions among the three types of anatomical correlations. Formally, the edge classifier \(f\) operates on the latent variable \(\mathbf{z}\) sampled from the posterior distribution \(q_\phi(\mathbf{z}|\mathbf{x}, \mathbf{y})\), where \(\mathbf{x}\) represents the concatenated edge features and \(\mathbf{y}\) denotes the edge label embeddings.

The objective of the CVAE is extended to include both the standard VAE loss and a classification loss for the edge classifier. This combined loss function is expressed as:

\[
\mathcal{L}_{\text{CVAE}} = \mathcal{L}_{\text{reconstruction}} + \lambda \mathcal{L}_{\text{edge-class}}
\]
\[
\mathcal{L}_{\text{reconstruction}} = \mathbb{E}_{q_\phi(\mathbf{z}|\mathbf{x}, \mathbf{y})} \left[ -\log p_\theta(\mathbf{x}|\mathbf{z}, \mathbf{y}) \right]
\]

where:
\begin{itemize}
    \item \(-\log p_\theta(\mathbf{x}|\mathbf{z}, \mathbf{y})\) is the reconstruction loss, ensuring the input features are faithfully reconstructed.
    \item \(\mathcal{L}_{\text{edge-class}}\) is the classification loss for the edge classifier, guiding the latent space to encode meaningful anatomical edge correlation information.
    \item \(\lambda\) are weighting factors that balance the contributions of the respective loss components.
\end{itemize}

By combining reconstruction, regularization, and classification objectives, the CVAE effectively captures both the input structure and the semantic relationships among anatomical correlations in the latent space, enabling robust feature representation and meaningful latent variable generation.

\paragraph{Edge Attention Mechanism}

To further enhance the model's capacity to focus on the most informative inter-sensor relationships, we incorporate an attention mechanism over edges. The attention mechanism allows the model to adaptively adjust the contribution of different sensor correlations based on their relevance to the activity recognition task as shown in Figure~\ref{GNN-ADG-framework} Edge Attention Mechanism component.

\textbf{1. Compute Attention Scores}

Given the learned edge features \(\mathbf{z}\), the attention scores \( s_e \) for each edge \( e \) are computed using a linear transformation followed by a Leaky ReLU activation:

\[
s_e = \text{LeakyReLU}\left( \mathbf{w}^\top \mathbf{z}_e + b \right),
\]

where:
\begin{itemize}
    \item \( \mathbf{z}_e \in \mathbb{R}^{\text{latent\_dim}} \): learned edge feature vector of edge \( e \),
    \item \( \mathbf{w} \in \mathbb{R}^{\text{latent\_dim}} \): learnable weights of the linear transformation,
    \item \( b \in \mathbb{R} \): learnable bias term.
\end{itemize}

\textbf{2. Compute Attention Weights}

The attention weights \( \alpha_e \) are obtained by applying the softmax function over the attention scores across all edges:

\[
\alpha_e = \frac{\exp(s_e)}{\sum_{e'} \exp(s_{e'})},
\]

where the sum is over all edges \( e' \) in the graph.

\textbf{3. Compute Weighted Edge Features}

The weighted edge features \(\tilde{\mathbf{Z}}\) are calculated by their corresponding attention weights: 

\[
\tilde{\mathbf{z}}_e = \alpha_e \cdot \mathbf{z}_e.
\]

The rationale for incorporating the attention mechanism is that not all anatomical correlations contribute equally to recognizing a particular activity. By assigning adaptive importance weights to edges, the model can prioritize the critical common relationships among users and focus on the most relevant inter-sensor interactions for each activity. This adaptability is crucial for handling domain shifts and user variability, as it allows the model to dynamically adjust its focus based on the input data.

\subsubsection{Node Feature Extractor}

Parallel to edge processing, the Node Feature Extractor operates on raw sensor signals, designed to identify position-invariant patterns from data. Its primary role is to extract localized features from time-series data collected from multiple body-worn sensors, such as those on the wrists, ankles, and body trunk. By ensuring consistent capture of fundamental low-level features across different sensor placements, it enhances robustness and generalizability in downstream processing. 

Let \(\mathbf{X} \in \mathbb{R}^{S \times T \times C}\) denote the input tensor, where:
\begin{itemize}
    \item \(S\): Number of body-worn sensors (nodes),
    \item \(T\): Time steps in the sensor signal,
    \item \(C\): Input channels (e.g., accelerometer and gyroscope \(x,y,z\) axes).
\end{itemize}

Acting as a generalized encoder, the node feature extractor processes signals uniformly across all sensor positions, learning shared representations that do not depend on specific sensor locations. This design promotes robust generalization, as the model captures common signal characteristics across different users.

To achieve this, we utilize a convolutional neural network (CNN) with two convolutional layers to extract these common features, as illustrated in Figure~\ref{GNN-ADG-framework} Node Feature Extractor component. The first convolutional layer applies filters to detect short-term temporal patterns, while the second convolutional layer builds on these extracted features to capture more complex structures. Both layers incorporate batch normalization, ReLU activation, and max-pooling, which help in dimensionality reduction and prevent overfitting. Mathematically, the node feature extractor is defined as:
\[
\mathbf{H} = f_{\text{NFE}}(\mathbf{X}; \theta),
\]
where:
\begin{itemize}
    \item \(\mathbf{H} \in \mathbb{R}^{S \times D}\): Output node features, with \(D\) as the learned feature dimension,
    \item \(\theta\): Shared CNN parameters (convolutional filters, biases, batch normalization terms),
    \item \(f_{\text{NFE}}\): Composition of convolution, activation, pooling, and flattening operations.
\end{itemize}

\subsubsection{Anatomical Correlation Knowledge Extractor}

To capture the interplay between sensor signals and their anatomical relationships, we fuse the node features with the weighted edge features using a \textbf{Anatomical Correlation Knowledge Extractor} (see Figure~\ref{GNN-ADG-framework}), which integrates weighted edge features and node features within the graph convolution mechanism \cite{gilmer2017neural}. This component is designed to capture complex and dynamic relationships between Anatomical Units (i.e., sensor positions), which is crucial for accurately modeling human activities involving coordinated movements across multiple body parts. The graph convolution is detailed in the following three sequential steps.

\begin{figure}[h!]
  \centering
  \includegraphics[width=0.45\columnwidth]{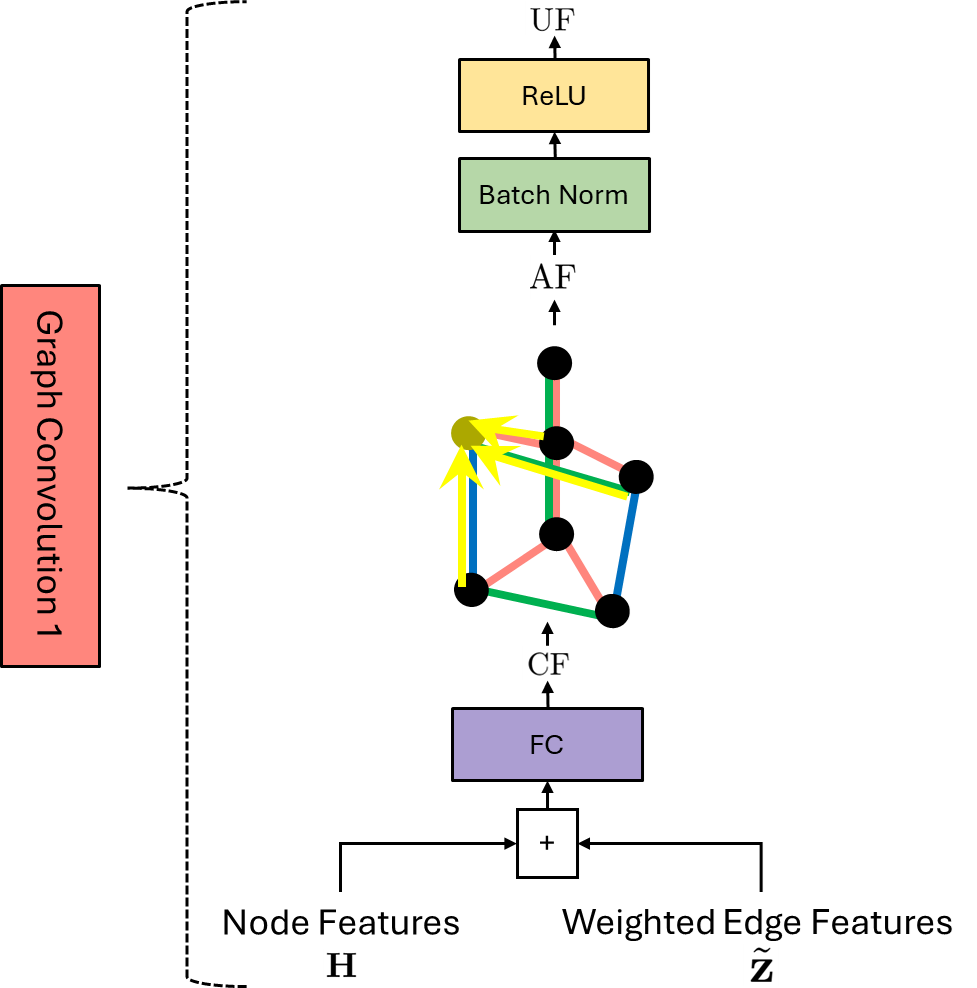}
  \caption{Edge-Enhanced Graph Convolution Layer. \label{GNN-framework}}
\end{figure}

For fusing the information from both node and edge features, we first compute a combined features (CF) that integrates the features of each node with its all corresponding connected edges' features. This transformation is defined as:
\begin{equation}
\text{CF} = \text{FC}\left( \left[ \mathbf{H} \, \| \, \tilde{\mathbf{Z}} \right] \right).
\end{equation}
where:
\begin{itemize}
    \item $\mathbf{H}$: Node features from previos step.
    \item $\tilde{\mathbf{Z}}$: Weighted edge features from Variational Edge Feature Extractor component.
    \item $\left[  \, \| \,  \right]$: Concatenation operation of the node feature and its connected edges features.
    \item $\text{FC}(\cdot)$: A fully connected layer that unify and compress the concatenated features into a combined feature vector.
\end{itemize}

Next, for each node $i$, the combined features from all its neighboring nodes are aggregated, which forms the node's aggregated features (AF). This is achieved by taking an average over the set of neighbors $\mathcal{N}(i)$:
\begin{equation}
\text{AF}_i = \frac{1}{|\mathcal{N}(i)|} \sum_{j \in \mathcal{N}(i)} \text{CF}_j.
\end{equation}
where:
\begin{itemize}
    \item $\mathcal{N}(i)$: The set of neighboring nodes of node $i$.
    \item $|\mathcal{N}(i)|$: The number of neighbors (cardinality of the set) of node $i$.
\end{itemize}

For example, in Figure~\ref{GNN-framework}, three yellow edges converge on the dark yellow node, showing the direction of the aggregation operation.

Finally, the aggregated features are passed through batch normalization to stabilize and scale the activations, followed by a non-linear activation function to introduce complexity and enable the model to learn intricate patterns. This process forms the final updated features (UF), which represent the fused knowledge of nodes and edges, and the final node representations in this graph convolution:
\begin{equation}
\text{UF} = \text{ReLU}\Bigl( \text{BN}\bigl( \text{AF} \bigr) \Bigr).
\end{equation}
where:
\begin{itemize}
    \item $\text{BN}(\cdot)$: Batch normalization that standardizes the aggregated message to stabilize and accelerate training.
    \item $\text{ReLU}(\cdot)$: ReLU non-linear activation function that introduces non-linearity into the updated node features.
\end{itemize}

Next, the second graph convolution layer operates in the same manner as the first layer, with the key difference being that its input consists of the UF generated by the first graph convolution layer. Following this, the output is passed through a fully connected layer for further transformation, which is then subjected to batch normalization and a ReLU activation function to enhance non-linearity and stabilize learning. Subsequently, global mean pooling is applied to aggregate the node-level features into a fixed-size representation, resulting in the final Graph Embedding G.

This Graph Embedding G encapsulates comprehensive information by integrating the node features extracted from the Node Feature Extractor component with the weighted edge features derived from the Variational Edge Feature Extractor component. Through this integration, the Anatomical Correlation Knowledge Extractor effectively captures intricate and variable relationships among sensor positions, leveraging both local node characteristics and global structural dependencies within the graph.

\subsubsection{Source Users Discriminator}

This component ensures that the extracted features remain domain-invariant, allowing them to generalize effectively across different users, as illustrated in Figure~\ref{GNN-ADG-framework} Source Users Discriminator component. To achieve this, a cyclic adversarial training strategy is adopted, alternating between \textit{discrimination} and \textit{confusion} phases. Let 
M denote the number of epochs allocated to each phase, forming a 2M-epoch training cycle that iteratively refines user-agnostic representations. Gradient Reversal Layer (GRL) is employed for adversarial learning. The GRL plays a crucial role by inverting the gradient direction during training, enabling the model to develop user-independent representations.

Mathematically, the GRL can be described as:
\[
\text{GRL}(x) = x, \quad \frac{d}{dx} \text{GRL}(x) = -I,
\]
where \( I \) is the identity matrix. This inversion ensures that the gradients are reversed during backpropagation, facilitating adversarial learning.

\textbf{Phase 1: Discrimination Phase (Epochs \( 1 \)--\( M \)).}  
In this phase, the Source Users Discriminator \( D \) is trained to classify user domains using the graph embedding \( G \), which integrates node and edge features from the Anatomical Correlation Knowledge Extractor. The discriminator minimizes the cross-entropy loss:
\[
\mathcal{L}_D = -\mathbb{E}_{G \sim \mathcal{G}} \left[ \sum_{u=1}^{U} y_u \log D(G) \right],
\]
where \( y_u \) is the one-hot encoded label for user \( u \), and \( U \) is the total number of users.  

\textbf{Phase 2: Confusion Phase (Epochs \( M+1 \)--\( 2M \)).}  
Here, the GRL is activated to reverse gradients from \( D \), forcing the model to generate user-invariant representations. The loss aims to maximize the discriminator’s error: \(-\mathcal{L}_D\).
The combined objective function alternates between phases:
\[
\mathcal{L}_{discriminator} = \mathbb{I}_{\text{Phase1}} \cdot \mathcal{L}_D + \mathbb{I}_{\text{Phase2}} \cdot (-\zeta \mathcal{L}_D),
\]
where:
\begin{itemize}
    \item \( \mathbb{I}_{\text{Phase1}} = \begin{cases} 1 & \text{for Epochs } 1\text{--}M \\ 0 & \text{otherwise} \end{cases} \)
    \item \( \mathbb{I}_{\text{Phase2}} = \begin{cases} 1 & \text{for Epochs } M+1\text{--}2M \\ 0 & \text{otherwise} \end{cases} \)
    \item \( \zeta \): Hyperparameter controlling adversarial strength
\end{itemize}

This cyclic strategy creates a push-and-pull dynamic: the model sharpens its discriminative capability in Phase 1, while disrupts user-specific cues in Phase 2. Over multiple \( 2M \)-epoch cycles, the model converges to an equilibrium where features are robustly domain-invariant, eliminating the need for user-specific calibration.  

\subsubsection{Activities Classifier}

The Activities Classifier is responsible for performing the HAR task, as depicted in Figure~\ref{GNN-ADG-framework} Activities Classifier component. It processes the graph embedding \( G \) to identify the specific activity being performed, such as walking, running, or sitting. The classifier maps these features to their corresponding activity labels, ensuring both generalization across users and high recognition accuracy.

The classifier consists of fully connected layers followed by ReLU activations and a softmax layer. Given the input graph embedding \( G \), the classifier outputs a probability distribution \( \mathbf{p}(G) \) over the activity classes. The classifier is trained using the cross-entropy loss function:
\[
\mathcal{L}_{\text{HAR}} = -\sum_{i=1}^N \sum_{c=1}^A y_{i,c} \log(p_{i,c}(G_i)),
\]
where:
\begin{itemize}
    \item \( N \): Number of training samples,
    \item \( A \): Number of activity classes,
    \item \( y_{i,c} \): Ground truth label (one-hot encoded) for sample \( i \) and class \( c \),
    \item \( p_{i,c}(G_i) \): Predicted probability for sample \( i \) and class \( c \), computed from the graph embedding \( G_i \).
\end{itemize}

By focusing exclusively on HAR, the classifier ensures that the representations learned by preceding components are effectively utilized for activity prediction. Since it leverages features that capture common anatomical correlations across users, the classifier achieves reliable and consistent activity recognition across different individuals.

\subsubsection{Overall Interaction of Components}
The EEG-ADG framework is designed to operate synnergy, where each components contributes to the extraction of robust, domain-invariant features that enable accurate cross-user HAR with common anatomical correlation knowledge. The process begins with the Variational Edge Feature Extractor component. In this stage, structured edge features are first generated and augmented with Gaussian noise to introduce realistic variability. These features, in combination with semantic edge labels that encode different anatomical correlations, are transformed into a latent space via a CVAE. This latent space is designed to highlight shared anatomical patterns across users, providing a reliable and generalizable representation of the inter-sensor relationships.

Following this, an attention mechanism is applied to the latent edge representations. This mechanism computes attention scores for each edge and selectively emphasizes those edges that are most informative. By prioritizing robust inter-sensor relationships, the attention mechanism ensures that noisy or less relevant connections exert a diminished influence on the overall representation. Simultaneously, node features are extracted directly from raw sensor data using Node Feature Extractor. These features capture local temporal dynamics and are inherently position-invariant. The Anatomical Correlation Knowledge Extractor then fuses the refined edge features with these node features. This fusion process results in a comprehensive representation that reflects both individual sensor behaviors and their interdependencies.

To further ensure that the learned features are user-invariant, a Source Users Discriminator is incorporated into the framework. Through adversarial learning—facilitated by a GRL—this discriminator forces the network to remove any user-specific cues from the combined representation. Finally, the Activities Classifier leverages these domain-invariant features to perform HAR. By utilizing representations that integrate both local sensor information and robust anatomical correlations, the classifier is able to accurately recognize activities across diverse users. Overall, the careful integration and interaction of these components allow EEG-ADG to effectively mitigate cross-user variability, delivering robust and generalizable performance in human activity recognition tasks.

\section{Experimental Setup}
\label{sec:experimental_setup}

To evaluate the effectiveness of our proposed \textbf{EEG-ADG} method for cross-user activity recognition, we conducted experiments using two widely recognized benchmark datasets: \textit{OPPORTUNITY} (OPPT) and \textit{Daily and Sports Activities Dataset} (DSADS). These datasets encompass a diverse range of activities and sensor setups, making them well-suited for testing the generalization capabilities of our approach. Table~\ref{tab_datasets_info} provides a summary of the subject clusters, shared activities across users, and sensor placements for both datasets.

\begin{table}[h!]
\caption{Two sensor-based HAR datasets information.}
\label{tab_datasets_info}
\centering
\scriptsize 
\resizebox{0.5\columnwidth}{!}{%
\begin{tabular}{|p{1.0cm}|p{1.5cm}|p{2.6cm}|}
\hline
\textbf{Dataset} & \textbf{Domains} & \textbf{Sensor Positions} \\ \hline
OPPT & \begin{tabular}[c]{@{}l@{}}A = [S1], \\
B = [S2],\\
C = [S3],\\
D = [S4]\end{tabular} &\begin{tabular}[c]{@{}l@{}} 1: Back,\\ 2: Right Upper Arm,\\ 3: Right Lower Arm,\\ 4: Left Upper Arm,\\ 5: Left Lower Arm \end{tabular}\\ \hline

DSADS & \begin{tabular}[c]{@{}l@{}}A = [1,2],\\
B = [3,4],\\
C = [5,6],\\
D = [7,8]\end{tabular} & \begin{tabular}[c]{@{}l@{}}1: Torso,\\ 2: Right Arm,\\ 3: Left Arm,\\ 4: Right Leg,\\ 5: Left Leg\end{tabular} \\ \hline
\end{tabular}%
}
\end{table}

\subsection{Datasets and Experimental Setup}

The \textbf{OPPT} dataset~\cite{chavarriaga2013opportunity} captures data from multiple body-worn sensors during daily living activities. It includes recordings from four subjects performing tasks such as walking, standing, sitting, and interacting with objects. This dataset presents significant challenges due to the inclusion of fine-grained activities involving highly similar motions, such as opening and closing various drawers (e.g., Drawer 1, 2, 3) or doors. Differentiating these actions relies on detecting subtle variations in sensor data corresponding to the specific location or layer being accessed. The dataset offers rich sensory inputs from accelerometers, gyroscopes, and magnetometers attached to different body parts.

The \textbf{DSADS} dataset~\cite{barshan2014recognizing} includes data from eight subjects performing 19 activities, spanning both daily and sports-related tasks. Each participant wore inertial measurement units on five body locations, generating comprehensive motion data. Unlike OPPT, the activities in DSADS, such as cycling, rowing, and playing basketball, generally exhibit more distinct motion patterns, which makes classification relatively less ambiguous.

We utilized a leave-one-subject-out cross-validation protocol to simulate a cross-user activity recognition scenario. For each dataset, the model was trained using data from all subject clusters except one, and the left-out cluster was used for testing. This approach ensures evaluation on entirely unseen users, providing a robust measure of the model's generalization performance.

\subsection{Cross-user HAR performance results}

To evaluate the effectiveness of our proposed \textbf{EEG-ADG} method, we compare it against the following state-of-the-art domain generalization approaches:

\begin{itemize} \item \textbf{ERM} \cite{zhang2018mixup}: Empirical Risk Minimization (ERM) is a fundamental approach that minimizes the average loss across all training samples. It serves as a baseline for domain generalization, focusing solely on learning from the data without explicitly addressing domain-specific differences.

\item \textbf{ANDMask}~\cite{parascandolo2021learning}: Aims to learn invariant features by enforcing updates that improve performance across all training domains simultaneously. By applying a logical AND operation, it emphasizes invariance while avoiding the overfitting of spurious features, making it particularly effective for tasks with clear distinctions between invariant and domain-specific features.

\item \textbf{RSC}~\cite{huang2020self}: Representation Self-Challenging (RSC) enhances cross-domain generalization by suppressing dominant features during training, encouraging the network to utilize alternative label-related features. This method is simple, effective, and requires no prior knowledge of unseen domains or additional parameters.

\item \textbf{DIFEX}~\cite{lu2022domaininvariant}: Domain-Invariant Feature Exploration (DIFEX) enhances generalization by integrating internal and mutual invariance. Internal invariance emphasizes intrinsic features within a domain, while mutual invariance identifies transferable features across domains.

\item \textbf{AdaRNN}~\cite{du2021adarnn}: Tackles the Temporal Covariate Shift issue in time-series data, where distribution changes over time reduce forecasting accuracy. AdaRNN uses two core algorithms: Temporal Distribution Characterization, which captures distribution trends, and Temporal Distribution Matching, which aligns distributions, ensuring adaptability and improved generalization.
\end{itemize}

\begin{table}[h!]
\caption{Two sensor-based HAR datasets Common Activities.}
\label{tab_datasets_info_common_activities}
\centering
\scriptsize
\resizebox{0.7\columnwidth}{!}{%
\begin{tabular}{|p{1.0cm}|p{8.0cm}|}
\hline
\textbf{Dataset} & \textbf{Common Activities} \\ \hline
OPPT & 
\begin{tabular}[t]{@{}l@{}}
1: Open Door 1, 2: Open Door 2, 3: Close Door 1,\\
4: Close Door 2, 5: Open Fridge, 6: Close Fridge,\\
7: Open Dishwasher, 8: Close Dishwasher, 9: Open Drawer 1,\\
10: Close Drawer 1, 11: Open Drawer 2, 12: Close Drawer 2,\\
13: Open Drawer 3, 14: Close Drawer 3, 15: Clean Table,\\
16: Drink From Cup, 17: Toggle Switch
\end{tabular} \\ \hline
DSADS & 
\begin{tabular}[t]{@{}l@{}}
1: Sitting, 2: Standing, 3: Lying On Back, \\
4: Lying On Right, 5: Ascending Stairs, \\
6: Descending Stairs, 7: Standing In Elevator Still, \\
8: Moving Around In Elevator, \\
9: Walking In Parking Lot, \\
10: Walking On Treadmill In Flat, \\
11: Walking On Treadmill Inclined Positions, \\
12: Running On Treadmill In Flat, \\
13: Exercising On Stepper, \\
14: Exercising On Cross Trainer, \\
15: Cycling On Exercise Bike In Horizontal Positions, \\
16: Cycling On Exercise Bike In Vertical Positions, \\
17: Rowing, 18: Jumping, 19: Playing Basketball
\end{tabular} \\ \hline
\end{tabular}%
}
\end{table}

These methods were chosen for their relevance and effectiveness in domain generalization tasks. To assess the effectiveness of our proposed approach, we conducted extensive experiments on two widely used benchmark datasets: DSADS and OPPT. Our model was compared with several state-of-the-art domain generalization techniques, including ERM, RSC, ANDMask, AdaRNN, and DIFEX. These comparisons show the advantages of EGG-ADG, which integrates edge-enhanced GNN with adversarial learning to improve generalization across users. By leveraging anatomical correlations between different body parts, EGG-ADG facilitates information fusion between biomechanical principles and machine learning methodologies, enhancing cross-user HAR performance.

\begin{figure*}[h!] \centering \includegraphics[width=0.8\textwidth]{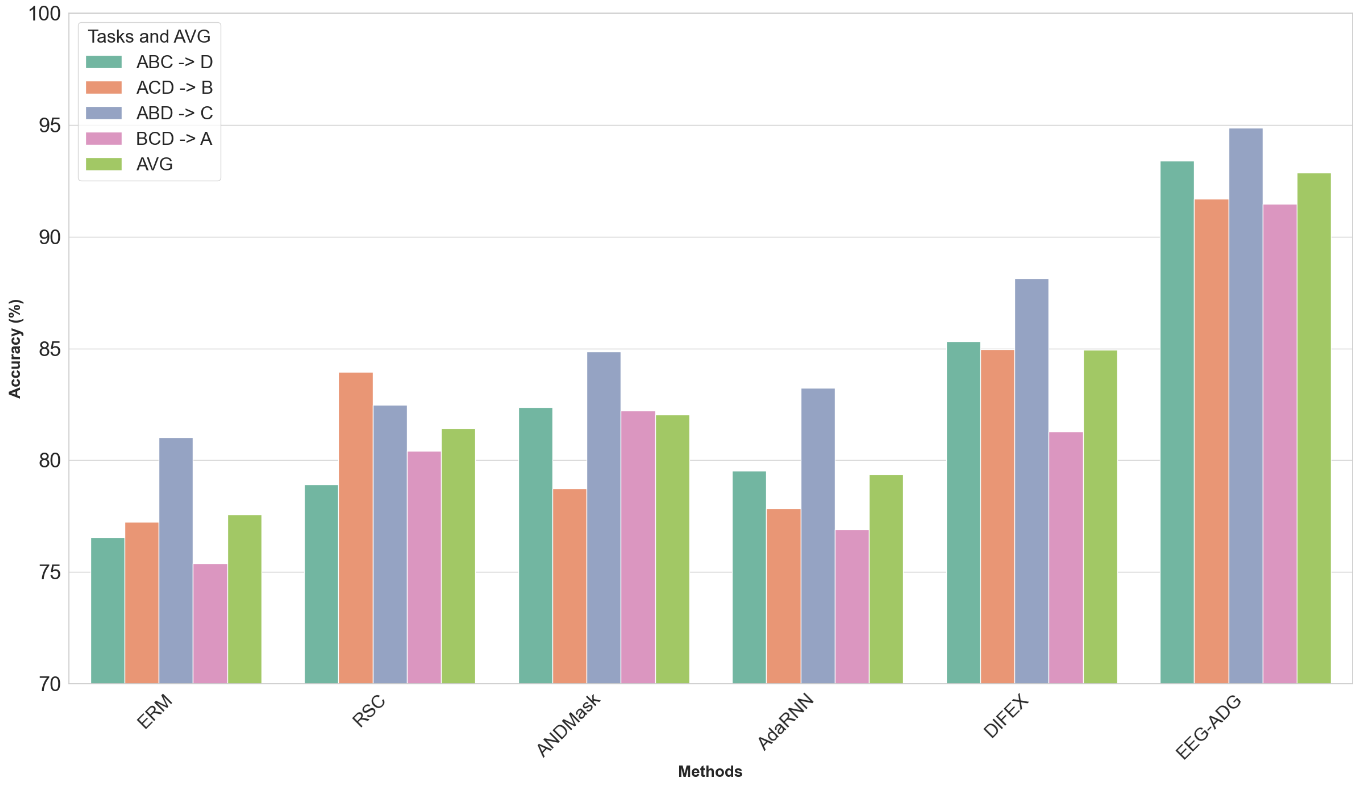} \caption{DSADS Dataset Results.\label{DSADS_Dataset_Result}} \end{figure*}

\begin{table}[ht]
\centering
\caption{Comparison of methods with average accuracy and standard deviation values on DSADS dataset.}
\begin{tabular}{lcc}
\toprule
\textbf{Method} & \textbf{Average} & \textbf{Standard deviation} \\
\midrule
ERM & 77.56 & 2.11 \\
RSC & 81.44 & 1.92 \\
ANDMask & 82.05 & 2.18 \\
AdaRNN & 79.38 & 2.41 \\
DIFEX & 84.93 & 2.44 \\
EGG-ADG & \textbf{92.85} & \textbf{1.38} \\
\bottomrule
\end{tabular}
\label{tab:comparison_updated_DSADS}
\end{table}

\textbf{DSADS Dataset.}
The DSADS dataset encompasses a range of activities, including both static postures (e.g., sitting, standing) and dynamic movements (e.g., walking, running) as shown in Table~\ref{tab_datasets_info_common_activities}. The dataset is captured using sensors positioned on the torso, arms, and legs to record full-body motion dynamics. Key observations from Table~\ref{tab:comparison_updated_DSADS} and Figure~\ref{DSADS_Dataset_Result} include:
\begin{itemize}
    \item Among the baseline methods, \textbf{DIFEX} achieves the highest accuracy (84.93\%) by promoting diversity and aligning domain-invariant features. However, it does not explicitly account for the interdependencies between various anatomical parts, which limits its effectiveness in complex activities like "rowing." Rowing is the activity where interdependencies between body parts are critical. It involves a sequence of movements where the legs, arms, and torso work in unison. The legs push, the torso leans back, and the arms pull the oar in a synchronized manner. Without modeling these interdependencies, DIFEX might struggle to recognize the activity accurately. For example, it might detect the arm movements but miss the contribution of the legs and torso, leading to misclassification or reduced accuracy.

    \item \textbf{EGG-ADG} surpasses all other methods, achieving the highest average accuracy (92.85\%). By incorporating three types of anatomical correlation knowledge-Interconnected Units, Analogous Units and Lateral Units, it effectively captures both local dependencies and global biomechanical structures. This is particularly beneficial in activities requiring full-body coordination. For instance, during "walking," Interconnected Units correlations between the upper and lower legs are crucial, whereas in "cycling," Analogous Units correlations between both legs facilitates better generalization.
\end{itemize}

\begin{figure*}[h!] \centering \includegraphics[width=0.8\textwidth]{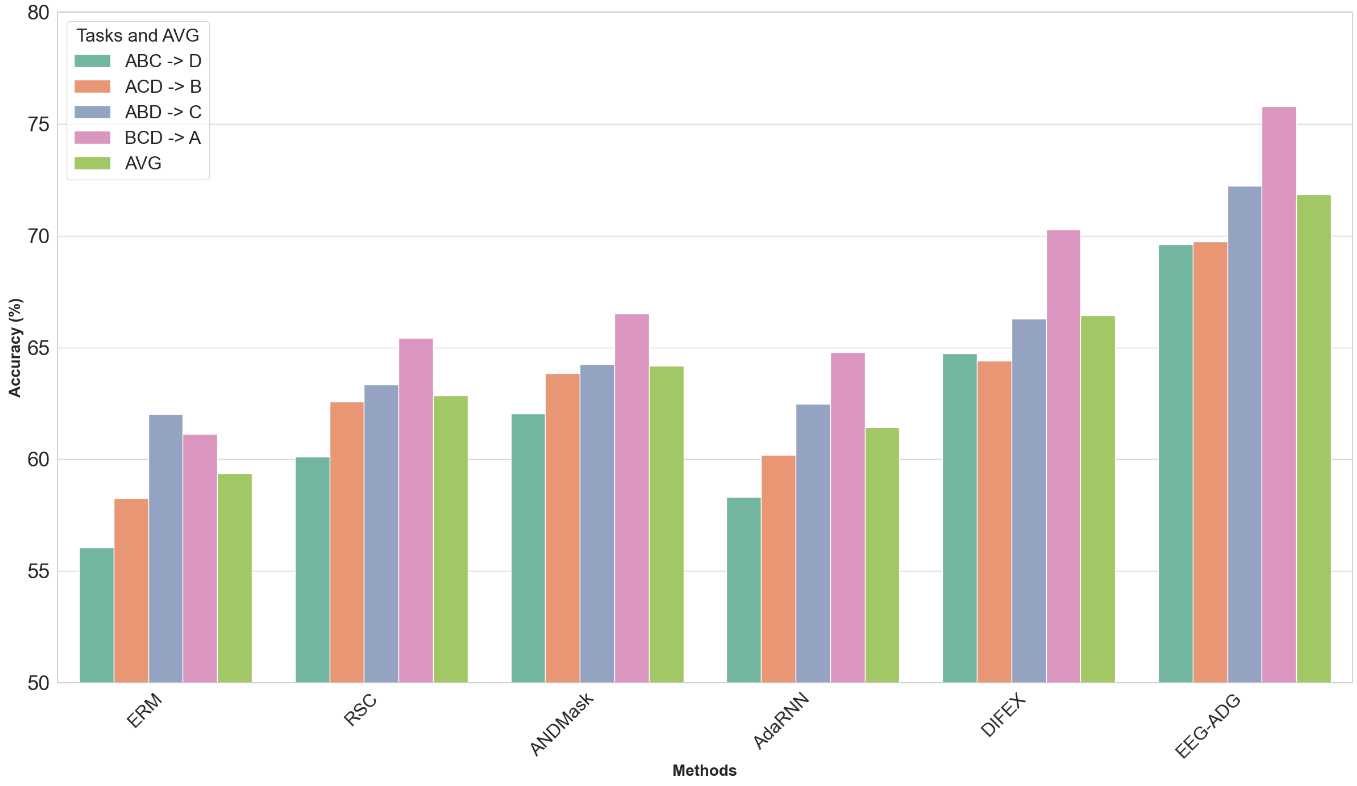} \caption{OPPT Dataset Results.\label{OPPT_Dataset_Result}} \end{figure*}

\begin{table}[h!]
\centering
\caption{Comparison of methods with average accuracy and standard deviation values on OPPT dataset.}
\begin{tabular}{lcc}
\toprule
\textbf{Method} & \textbf{Average} & \textbf{Standard deviation} \\
\midrule
ERM & 59.37 & 2.37 \\
RSC & 62.89 & 1.90 \\
ANDMask & 64.17 & \textbf{1.59} \\
AdaRNN & 61.44 & 2.43 \\
DIFEX & 66.43 & 2.33 \\
EGG-ADG & \textbf{71.84} & 2.51 \\
\bottomrule
\end{tabular}
\label{tab:comparison_OPPT}
\end{table}

\textbf{OPPT Dataset.}
The OPPT dataset focuses on upper-body movements, such as opening/closing doors, toggling switches, and cleaning tasks. Sensors are placed on the back, upper arms, and lower arms to capture localized upper-body movements. Insights drawn from Table~\ref{tab:comparison_OPPT} and Figure~\ref{OPPT_Dataset_Result} include:
\begin{itemize}
    \item Among the baseline methods, \textbf{DIFEX} achieves the highest accuracy (66.43\%), demonstrating its ability to generalize across users. However, it does not explicitly model localized correlations between adjacent body parts, which are crucial for OPPT tasks. In OPPT tasks, adjacent body parts (e.g., upper arms and lower arms, arms and torso) often work in a coordinated manner. These localized correlations are essential because upper-body activities involve precise, interdependent motions. For example, opening a door requires the lower arm (hand) to pull the handle while the upper arm (elbow and shoulder) positions the hand. The relationship between adjacent body parts provides context. Accurately interpreting this data requires modeling how these body parts interact.

    \item \textbf{EGG-ADG} achieves the best performance (71.84\%), outperforming all baseline models. It captures both localized and global biomechanical relationships. Localized relationships refer to the interactions between adjacent or closely connected body parts. These relationships are critical for understanding fine-grained, coordinated movements. For instance, the wrist rotates to align the hand with the drawer handle, while the hand adjusts its grip strength to pull the drawer open. This interaction between the wrist and hand is the localized relationship. Global relationships refer to the interactions between distant or non-adjacent body parts that contribute to the overall activity. These relationships provide a holistic understanding of the movement. For example, the torso leans forward or backward to maintain balance while the arms perform the pulling motion. The torso's movement influences the arms' range of motion, demonstrating a global relationship. As for the activity of "opening a drawer," Lateral Units correlations between the upper and lower arms, coupled with Analogous Units relationships (e.g., left vs. right arm), enhance generalization across users with more robust classification result.
\end{itemize}

While baseline methods such as ERM, RSC, ANDMask, AdaRNN, and DIFEX have demonstrated effectiveness in domain generalization, they do not explicitly utilize the biomechanical relationships among anatomical parts. ERM processes all features equally, missing essential spatial and functional correlations necessary for activities like "cycling" or "cleaning a table". RSC removes dominant features to improve generalization but does not capitalize on rich interdependencies between sensors. DIFEX aligns domain-invariant features but lacks the capacity to model hierarchical and coordinated relationships among body parts.

In contrast, EGG-ADG incorporates three key anatomical relationships—Interconnected Units, Analogous Units, and Lateral Units—into a unified graph structure. This structure enables the model to capture both local and global biomechanical patterns. Interconnected Units correlations emphasize immediate interactions between adjacent body parts, such as the coordination between upper and lower arms in OPPT's "cleaning a table." Analogous Units correlations model symmetrical or synchronized movements, which is crucial for DSADS activities like "cycling" and "rowing." Lateral Units correlations captures side-specific coordination, which plays a critical role in tasks like "opening a door," where distinguishing between active and stabilizing limbs is necessary. By integrating these relationships into a single framework, EGG-ADG effectively models inter-sensor dynamics, enhancing generalization across users and activities.

Moreover, the EGG-ADG employs several innovative components (i.e. Variational Edge Feature Extractor, Anatomical Correlation Knowledge Extractor, Source Users Discriminator) that individually and collectively address critical challenges in cross-user domain generalization for HAR. First, the CVAE in Variational Edge Feature Extractor is designed to probabilistically model edge features, enabling the model to account for user variability while preserving shared biomechanical patterns across individuals. By introducing a latent representation of edge features, the CVAE captures the natural variations in inter-sensor relationships caused by differences in anatomy or movement styles. For example, users may differ in how they perform tasks like "rowing," with some relying more on their arms while others engage their torso. The probabilistic nature of the CVAE ensures that the model learns a flexible representation that can generalize to unseen users by focusing on the underlying biomechanical structure rather than user-specific nuances. 

Second, for further enhance the model’s ability to focus on the most informative inter-sensor relationships, an attention mechanism in Variational Edge Feature Extractor is applied over edges. This mechanism adaptively adjusts the contribution of different sensor correlations based on their relevance to the activity being recognized. By assigning higher weights to the most relevant relationships, the attention mechanism improves the model’s robustness to user variability, dynamically adjusting its focus based on the input data.

Third, adversarial learning in Source Users Discriminator is adopted in EEG-ADG facilitated by a GRL, which is a crucial component for extracting user-invariant features. The GRL works by reversing the gradients of the domain discriminator during confusion phase, enabling the model to learn representations that confuse the domain discriminator. This setup encourages the model to produce features that are indistinguishable across different user domains, effectively making the features domain-invariant.

Finally, the Anatomical Correlation Knowledge Extractor dynamically captures anatomical correlations through the fusion of node features and weighted edge features, enabling the model to prioritize transferable patterns rooted in human movement physiology. By aggregating neighborhood interactions and normalizing structural representations, the extractor reduces sensitivity to user-specific variations in sensor placement or motion execution while emphasizing invariant biomechanical principles—such as joint kinematics and multi-limb coordination. The resulting graph embeddings encode both localized sensor dynamics and global anatomical topology, allowing the model to generalize to unseen users by leveraging universal biomechanical constraints rather than overfitting to superficial statistical artifacts.

EEG-ADG bridges domain knowledge (e.g., predefined anatomical connections) with learned data-driven adjustments, ensuring robustness to real-world variability.  Ultimately, this structured integration of anatomical knowledge enhances cross-domain adaptability, enabling reliable human activity recognition even under individual movement styles. The results across both datasets highlight the unique strengths of EEG-ADG. It combines GNN with adversarial learning to achieve domain generalization across users. By leveraging common anatomical correlations, EEG-ADG captures the intricate interplay of biomechanical relationships, enabling robust cross-user HAR performance. EEG-ADG demonstrates the highest average accuracy on both datasets, providing an improvement over existing methods and highlighting the importance of integrating spatial, positional, and functional relationships.

\subsection{Model Interpretability Analysis}

\begin{figure}[h!]
    \centering
 \includegraphics[width=0.7\textwidth]{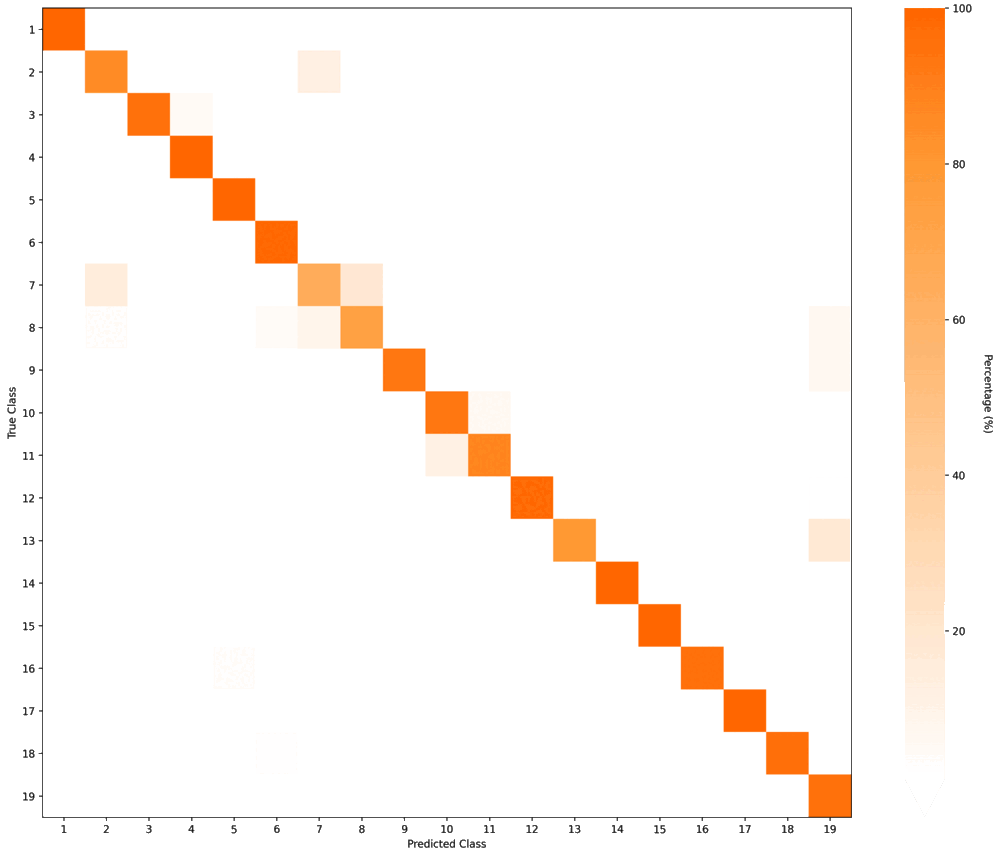}
    \caption{Confusion matrix for the DSADS dataset.\label{fig:dsads_cm}}
\end{figure}

\begin{figure}[h!]
    \centering
 \includegraphics[width=0.7\textwidth]{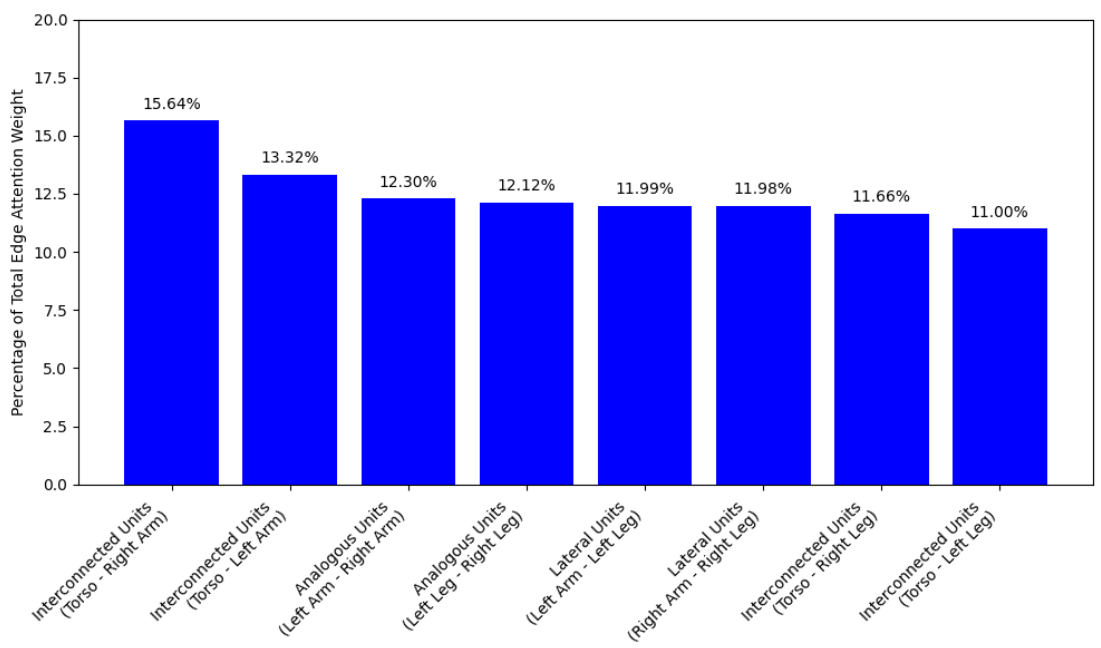}
    \caption{Edge attention weight for the DSADS dataset.\label{fig:dsads_attention}}
\end{figure}

\begin{figure}[h!]
    \centering
 \includegraphics[width=0.7\textwidth]{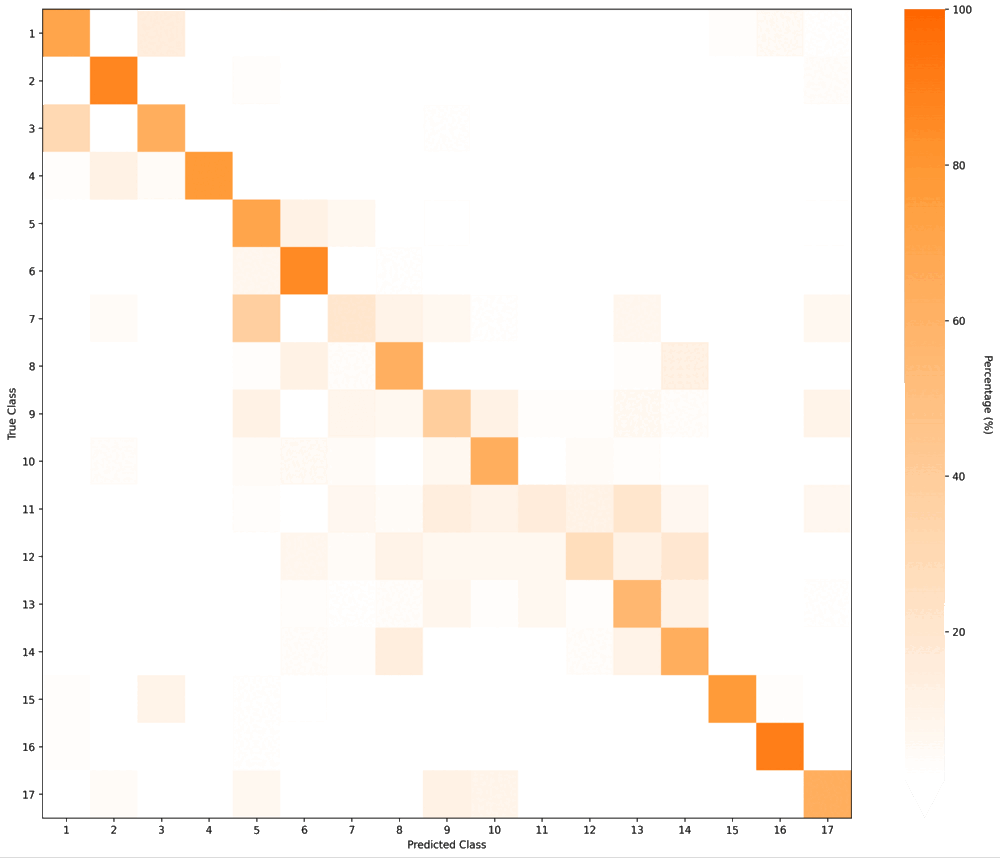}
    \caption{Confusion matrix for the OPPT dataset.\label{fig:oppt_cm}}
\end{figure}

\begin{figure}[h!]
    \centering
    \includegraphics[width=0.7\textwidth]{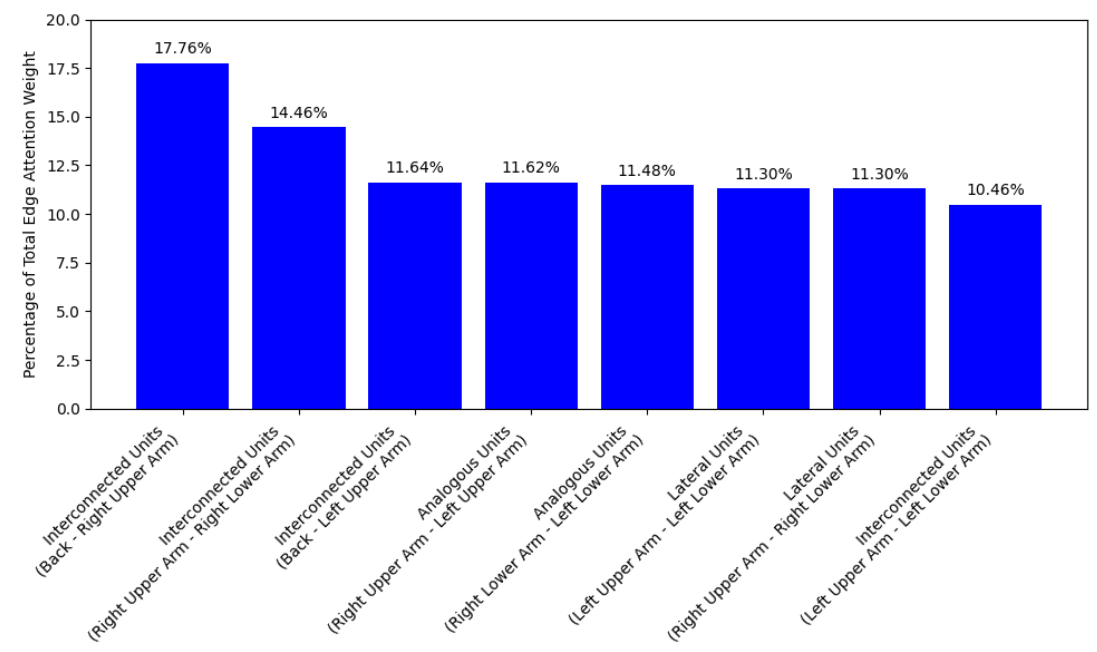}
    \caption{Edge attention weight for the OPPT dataset.\label{fig:oppt_attention}}
\end{figure}

To enhance the interpretability of the model’s predictions, we focus on the Edge Attention Mechanism, which plays a crucial role in understanding the contribution of different sensor connections within the learned graph structure. By analyzing the confusion matrix of the activity classes in conjunction with the attention weights assigned to edges, we can identify which sensor relationships are most influential in determining activity recognition outcomes. Figures~\ref{fig:dsads_cm} and~\ref{fig:oppt_cm} display the confusion matrices of the activity classification for the DSADS and OPPT datasets, respectively. Additionally, Figures~\ref{fig:dsads_attention} and~\ref{fig:oppt_attention} present the ranked edge-attention weights for the DSADS and OPPT datasets, respectively. The bar charts visualize the sensor connections assigned the highest attention weights by the learned model, revealing which anatomically correlated body-part relationships are prioritized as most influential for activity recognition tasks. This combined analysis allows us to better understand how the model leverages anatomical body-part relationships to make its classification, thereby improving the interpretability of its decision-making process.

For the DSADS dataset (Figure~\ref{fig:dsads_attention}), the top‐ranked edges primarily involve Interconnected Units between the torso and the arms (e.g., \textit{Torso--Right Arm} at $\approx 15.6\%$ and \textit{Torso--Left Arm} at $\approx 13.3\%$). We also observe moderately high attention weights for Analogous Units (e.g., \textit{Left Arm--Right Arm}) and Lateral Units (\textit{Left Arm--Right Leg}). These results suggest that for whole‐body activities in DSADS (e.g., running on a treadmill, ascending or descending stairs, rowing, etc.), the model relies on the coordination between upper‐body and trunk movements. In particular, \textit{torso--arm} edges appear critical for capturing pivotal gestures (arm swinging or posture changes), while Analogous Units (e.g., \textit{Left Arm--Right Arm}) can help distinguish symmetrical from asymmetrical patterns of movement. Since the activities of DSADS involving the entire body, it is consistent that torso--limb interactions and limb--limb synergies show the highest attention values.

Another insight from the edge attention weights in the DSADS dataset is that the upper-body interactions receive more attention compared to the lower-body, as evidenced by the top three edge weights all being associated with upper-body connections. This observation aligns with the relatively weaker classification performance for activities primarily driven by lower-body movements (i.e., leg-related actions). For instance, as shown in Figure~\ref{fig:dsads_cm}, activities such as 2: standing, 7: standing in an elevator, and 8: moving in an elevator are confused. Similarly, 10: walking on a treadmill on a flat surface and 11: walking on a treadmill on an inclined surface are misclassified. Additionally, 19: playing basketball is sometimes misclassified as 8: moving in an elevator or 9: walking in a parking lot, likely due to the subtle differences in leg movement patterns. These misclassifications highlight a limitation of the Edge Attention Mechanism, particularly in capturing fine-grained distinctions in lower-body dynamics. This suggests that while the mechanism effectively models upper-body coordination, it may require further refinement to better address activities dominated by lower-body movements.

For the OPPT dataset (Figure~\ref{fig:oppt_attention}), which emphasizes upper‐body actions (e.g., opening doors, using drawers, toggling switches), the highest attention weights are found in edges connecting the \emph{Back} with the \emph{Right Upper Arm} ($\approx 17.7\%$) and those between the \emph{Right Upper Arm--Right Lower Arm} ($\approx 14.4\%$). This reflects the crucial role of torso--arm coordination for fine‐grained manipulations that are primarily performed with the arms and hands. Edges classified as \emph{Analogous Units} (e.g., \textit{Right Upper Arm--Left Upper Arm}) or \emph{Lateral Units} still appear but with slightly lower attention, which suggests that \emph{unilateral} movements on one side typically dominate these functional tasks. 

Additionally, two lateral edges overlap with interconnected units: \textit{Right Upper Arm--Right Lower Arm} and \textit{Left Upper Arm--Left Lower Arm}. This duplication indicates that these edges are recognized under both \emph{Interconnected Units} and \emph{Lateral Units}, potentially reflecting their combined significance in upper‐body manipulations. The overlapping classification may lead the model to assign attention to these edges from multiple perspectives, underscoring their importance in tasks that require coordinated movements between upper and lower arm segments. Given OPPT's focus on object manipulation, it is logical that \textit{Back--Arm} edges receive higher attention, especially if the user is right‐hand dominant and thus more likely to engage the right arm in these tasks. The model places significant emphasis on \emph{Back--Arm} interconnections, capturing the essential arm movements required for precise interactions with objects.

Another insight from the edge attention weights in the OPPT dataset is that activities involving the opening and closing of the same object are frequently confused. For instance, as shown in Figure~\ref{fig:oppt_cm}, activities such as 1: open door 1 and 3: close door 1, 5: open fridge and 6: close fridge, and 9: open drawer and 10: close drawer are often misclassified. This suggests a limitation of the model in capturing temporal relationships, particularly when distinguishing between reverse sub-activity sequences. A more fine-grained data segmentation approach, such as using a smaller sliding window size, may be necessary to better capture these temporal nuances. Furthermore, the model struggles to differentiate between the same type of operation performed on different objects. For example, activities 5: open fridge, 7: open dishwasher 2, 9: open drawer, and 11: open drawer are all open operations, while 6: close fridge, 8: close dishwasher, and 12: close drawer are all close operations. Despite the Edge Attention Mechanism, such subtle differences remain challenging to distinguish. This highlights the need for additional mechanisms to better capture the contextual and temporal variations in these fine-grained activities.

Across both datasets, the highest‐attention edges correlate well with the specific body segments most involved in the respective activities. DSADS includes full‐body exercises, leading to greater emphasis on torso--limb and limb--limb interplay. In contrast, OPPT's tasks revolve around arm and hand usage, so the model assigns more weight to \textit{Back--Arm} interconnections. These patterns illustrate that the \emph{Edge Attention Mechanism} effectively identifies the sensor pairings most relevant to each dataset's typical activities, shows how the model integrates sensor data to make accurate predictions.

\subsection{Ablation Study Analysis}

\begin{figure*}[h!] 
\centering 
\includegraphics[width=0.6\textwidth]{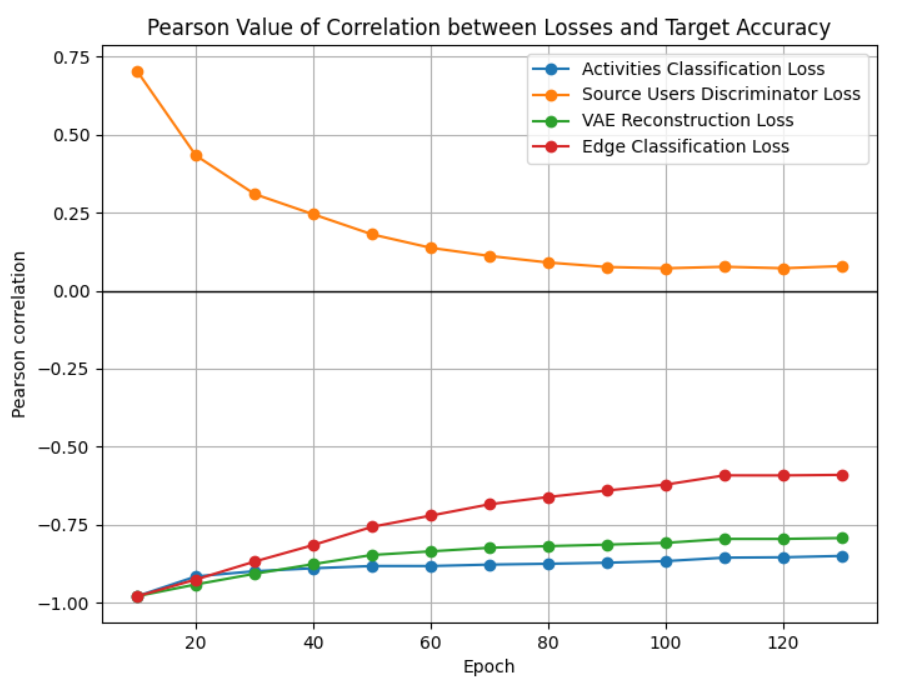} 
\caption{Pearson’s Correlation between Losses and Target Accuracy.\label{Ablation_Study_r}} 
\end{figure*}

\begin{figure*}[h!] 
\centering 
\includegraphics[width=0.6\textwidth]{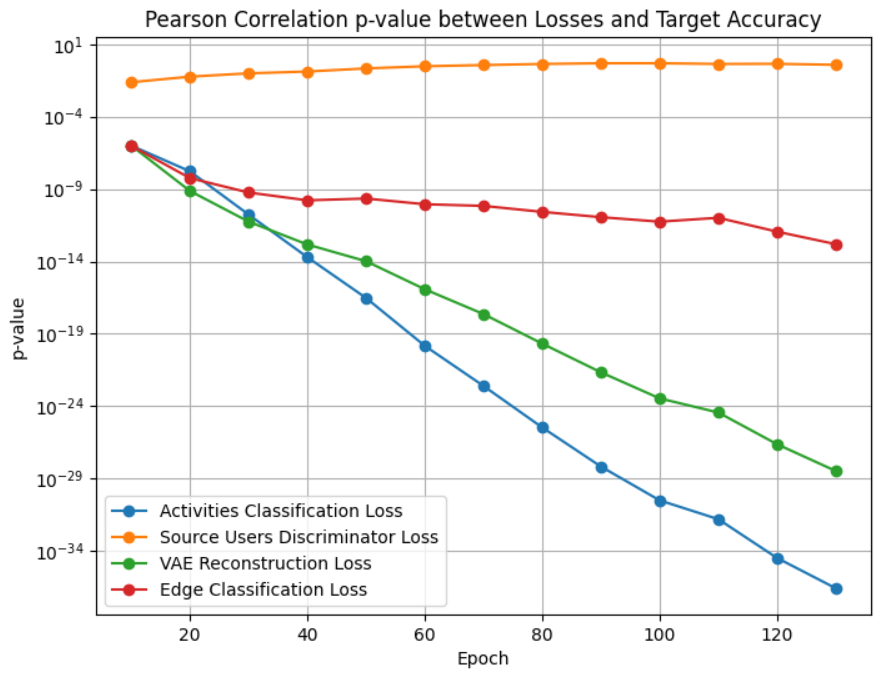} 
\caption{Pearson’s p-values for Losses versus Target Accuracy.\label{Ablation_Study_p}} 
\end{figure*}

The EEG-ADG method incorporates multiple loss functions, each designed to optimize specific objectives during training. To assess the impact of each loss component on the proposed method, we conduct a comprehensive correlation analysis between these losses and the target users accuracy. The results, displayed in Figure~\ref{Ablation_Study_r} and Figure~\ref{Ablation_Study_p}, include the Pearson correlation coefficients $r$ along with their statistical significance levels $p$‐value, providing insights into how each loss function influences the accuracy in the target domain.

To quantify the relationship between each loss term and the final target accuracy, we compute the Pearson correlation coefficient between each loss and the target users accuracy using progressively larger subsets of epochs (from 1--10, 1--20, \dots, up to 1--130). Figure~\ref{Ablation_Study_r} and Figure~\ref{Ablation_Study_p} summarize these results at 10‐epoch increments, including the corresponding $p$‐value for significance testing.

\begin{itemize}
    \item \textbf{Activities Classification Loss, VAE Reconstruction Loss, and Edge Classification Loss:} 
    All three exhibit \emph{strong negative correlation} with target accuracy throughout training, with correlation magnitudes typically above $\lvert r\rvert>0.80$ and $p$‐values far below $10^{-10}$.
    \begin{itemize}
        \item \emph{Early Epochs (1--10):} 
        Correlations approach $-0.98$ (all with $p\!\approx\!10^{-7}$), implying that even from the initial stages, minimizing these losses strongly coincides with increasing accuracy on the target domain.
        \item \emph{Mid to Late Epochs (1--130):} 
        Although the absolute magnitude of $r$ decreases slightly over time (e.g., from $-0.98$ down to about $-0.85$), the correlation remains highly significant ($p\!<\!10^{-20}$). These trends confirm that deeper minimization of activities classification, reconstruction, and edge‐classification losses continues to yield direct improvements in target accuracy as training progresses.
    \end{itemize}

    \item \textbf{Source‐Users Discriminator Loss:} 
In contrast, the source‐users discriminator (domain‐adversarial) loss acts as a regularization mechanism, encouraging data distribution alignment across source users. Empirically, it shows a moderate but \emph{statistically significant} positive correlation in the earliest epochs (e.g., $r=0.702$ for epochs 1--10 with $p=2.36\times10^{-2}$), indicating that forcing early domain alignment can help ``jump‐start'' better feature transfer. However, this correlation diminishes toward near zero, reflecting that once the model has sufficiently ``confused'' the source and target domains, \emph{further} accuracy gains rely more strongly on the downstream objectives (activities classification, VAE reconstruction, and edge classification). The small $p$‐value at early stages thus confirms the \emph{adversarial} synergy is initially influential for cross‐domain adaptation; thereafter, the domain‐confusion loss serves as a regularization rather than a direct driver of final performance gains.

\end{itemize}

In summary, these results demonstrate that minimizing the activities classification, VAE reconstruction, and edge classification losses remains critical to boosting target accuracy at all stages, and the adversarial source users discriminator objective exerts its most direct, measurable effect in the early epochs before its correlation with performance saturates.

\section{Conclusion}
\label{sec:conclusion}

In this paper, we introduced Edge-Enhanced Graph-Based Adversarial Domain Generalization (EEG-ADG), a novel framework that integrates anatomical correlation knowledge into graph-based adversarial learning for cross-user HAR. Our approach effectively addresses the challenge of cross-user variability by leveraging biomechanical priors and domain-invariant graph representations. Through Variational Edge Feature Extractor, EEG-ADG dynamically models sensor relationships, enabling robust and generalizable activity recognition. The incorporation of a GRL adversarial learning further enhances its domain generalization capability, ensuring the model remains effective for unseen users.

Experimental results on the OPPORTUNITY and DSADS datasets demonstrate the state-of-the-art performance of EEG-ADG, surpassing existing domain adaptation and generalization methods. By explicitly encoding Interconnected Units, Analogous Units, and Lateral Units, our framework bridges the gap between biomechanical principles and deep learning-based HAR, offering a scalable and adaptable solution for real-world applications in healthcare, sports analytics, and smart environments.

While EEG-ADG provides significant advancements in cross-user HAR, several directions for future research remain. Our current approach models anatomical correlations using a conventional graph structure, where each sensor is represented as a node and relationships as edges. A promising future direction is to explore hypergraphs, which can encode higher-order relationships among multiple sensors. This would enable more complex biomechanical interactions to be captured, further enhancing generalization capabilities. Moreover, real-world HAR systems often encounter missing or noisy sensor data. Future work could involve adaptive graph pruning or dynamic sensor selection, allowing the model to self-adjust based on available sensor inputs, improving robustness in practical deployments. By integrating these directions, we aim to further refine and extend EEG-ADG into a more adaptive, efficient, and generalizable HAR framework, capable of bridging biomechanical insights with deep learning advancements for broader real-world applications.

\bibliographystyle{elsarticle-num}
\bibliography{ref}

\end{document}